\documentclass[conference]{IEEEtran}
\IEEEoverridecommandlockouts
\usepackage{cite}
\usepackage{amsmath,amssymb,amsfonts}
\usepackage{algorithmic}
\usepackage{graphicx}
\usepackage{textcomp}
\usepackage{tabularx}
\usepackage{xcolor}
\usepackage{multirow}
\def\BibTeX{{\rm B\kern-.05em{\sc i\kern-.025em b}\kern-.08em
    T\kern-.1667em\lower.7ex\hbox{E}\kern-.125emX}}
\begin{document}

\title{Multi-modality action recognition based on dual feature shift in vehicle cabin monitoring\\}

\author{
	\IEEEauthorblockN{
		Dan Lin\IEEEauthorrefmark{1},
		Philip Hann Yung Lee\IEEEauthorrefmark{2},
        Yiming Li\IEEEauthorrefmark{2},
        Ruoyu Wang\IEEEauthorrefmark{2},
        Kim-Hui Yap\IEEEauthorrefmark{2}$^\ast$,
        Bingbing Li\IEEEauthorrefmark{3},
		and You Shing Ngim\IEEEauthorrefmark{3}}
	\IEEEauthorblockA{\IEEEauthorrefmark{1}Continental-NTU Corporate Lab, Nanyang Technological University, Singapore\\}
	\IEEEauthorblockA{\IEEEauthorrefmark{2}School of Electrical and Electronic Engineering, Nanyang Technological University, Singapore\\}
	\IEEEauthorblockA{\IEEEauthorrefmark{3}Continental Automotive Singapore Pte. Ltd., Singapore\\}
    $^{\ast}$\textit{Corresponding author: EKHYap@ntu.edu.sg}
}

\maketitle

\begin{abstract}
Driver Action Recognition (DAR) is crucial in vehicle cabin monitoring systems. In real-world applications, it is common for vehicle cabins to be equipped with cameras featuring different modalities. However, multi-modality fusion strategies for the DAR task within car cabins have rarely been studied. In this paper, we propose a novel yet efficient multi-modality driver action recognition method based on dual feature shift, named DFS. DFS first integrates complementary features across modalities by performing modality feature interaction.
Meanwhile, DFS achieves the neighbour feature propagation within single modalities, by feature shifting among temporal frames. To learn common patterns and improve model efficiency, DFS shares feature extracting stages among multiple modalities. Extensive experiments have been carried out to verify the effectiveness of the proposed DFS model on the Drive\&Act dataset. The results demonstrate that DFS achieves good performance and improves the efficiency of multi-modality driver action recognition.
% most existing DAR methods mainly focus on the single modality input. 

% we open the eyes from single-modality to multi-modality.

%pl In addition, we design to share the feature extracting stages among multiple modalities for model efficiency
% In addition,
 
% for model efficiency.
% with shared convolutional layers for training efficiency. design a dumbbell-shaped framework where multiple modalities

% accurate, as well as efficient, for driver action recognition.
% first conducts modality feature interactions by feature shifting among different modalities. 

% along the modality dimension.
% shifts the features along the temporal dimension within a single modality, thereby facilitating feature propagation between frames

% The abstract should appear at the top of the left-hand column of text, about
% 0.5 inch (12 mm) below the title area and no more than 3.125 inches (80 mm) in
% length.  Leave a 0.5 inch (12 mm) space between the end of the abstract and the
% beginning of the main text.  The abstract should contain about 100 to 150
% words, and should be identical to the abstract text submitted electronically
% along with the paper cover sheet.  All manuscripts must be in English, printed
% in black ink.
\end{abstract}
\section{Introduction}
\label{sec:intro}
The Driver Action Recognition (DAR) task involves automatically identifying drivers' secondary activities within the vehicle cabin during driving \cite{DBLP:journals/sensors/TrabelsiKDEB22}. DAR is crucial for enhancing driving safety and promoting efficient interactions between humans and vehicles.
Recently, significant progress on the DAR task has been achieved due to the advances in automation technologies and the application of deep learning methods \cite{DBLP:journals/itsm/RongHHLK22}. 
Several methods have been proposed for DAR, often building upon general human action recognition models that utilize 3D convolutional neural networks (CNNs) \cite{chen2023skateboardai} and vision transformers \cite{wang2023videomaev2}. 
Among them, the temporal shift module (TSM) provides an efficient solution by shifting features from neighbour frames \cite{DBLP:conf/iccv/LinGH19}.
However, existing research primarily concentrates on extracting spatial-temporal features to enhance DAR performance within single-modality input, such as
%pl Infrared Radiation (IR) Peng2022TransDARCTD
Infra-red (IR) video frames \cite{Wharton2021CoarseTA}.

%pl Features from the single-modality input can be insufficient to support accurate long-term action recognition. On the one hand, the car cabin environment is complex, with limited features available. As shown in Fig.~\ref{sample}, drivers' actions are performed by the same individual with only a portion of the body (such as the upper body) visible and often exhibit highly similar body part movements (such as eating and drinking). On the other hand, the lighting conditions within the cabin are unstable, depending on factors such as weather and transportation infrastructure, and can be affected by sub-optimal factors like sunlight variation. For example in Fig.~\ref{sample}, the RGB and IR frames can be in low brightness or exposed impacted by the weather. In light of these challenges, incorporating features from multiple modalities becomes crucial for effectively addressing the complex and difficult DAR task.

\begin{figure}[t]
% \vspace{2.5cm}
\centering
\includegraphics[width=0.45\textwidth,keepaspectratio]{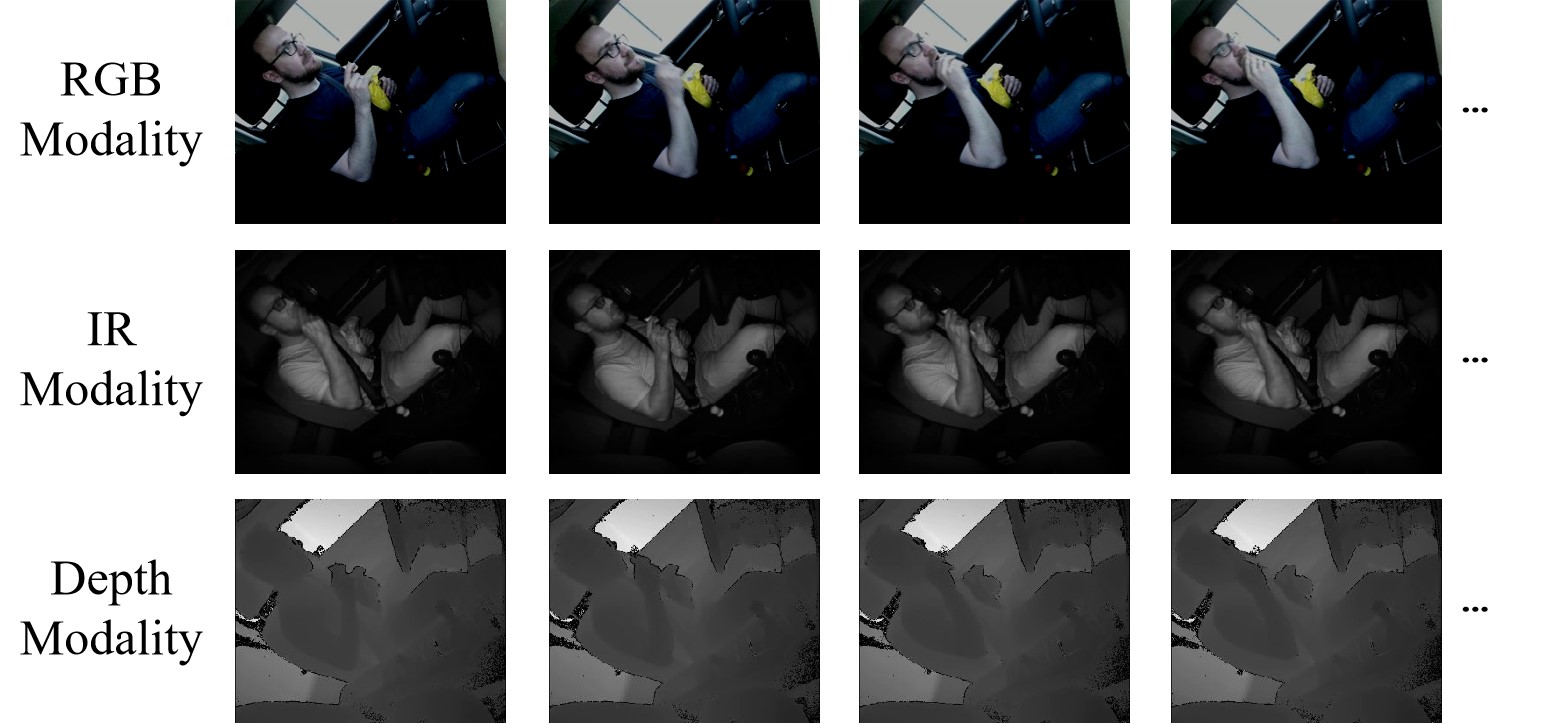}
\vspace{-3mm}
\caption{Sample frame sequences from different modalities for the action 'eating'. For each modality, drivers' actions are performed by the same individual with only a portion of the body visible and in unstable lighting conditions.}
\label{sample}
\vspace{-5mm}
\end{figure}

Features from a single-modality input may be insufficient to accurately support long-term action recognition. First, the car cabin environment is complex, with limited available features. As shown in Fig.~\ref{sample}, drivers' actions are performed by the same individual, with only a portion of the body (such as the upper body) visible. They often exhibit highly similar movements of body parts (such as eating and drinking). Second, the lighting conditions within the cabin are unstable, depending on factors such as weather and transportation infrastructure, and can be affected by sub-optimal factors like variations in sunlight. For example, in Fig~\ref{sample}, the RGB and IR frames can have low brightness or be exposed to weather-related conditions. Given these challenges, incorporating features from multiple modalities becomes crucial for effectively addressing the complex and demanding DAR task.

% of driver action recognition.
% 加个图片

\begin{figure*}[t]
% \vspace{2.5cm}
\centering
\includegraphics[width=0.9\textwidth,keepaspectratio]{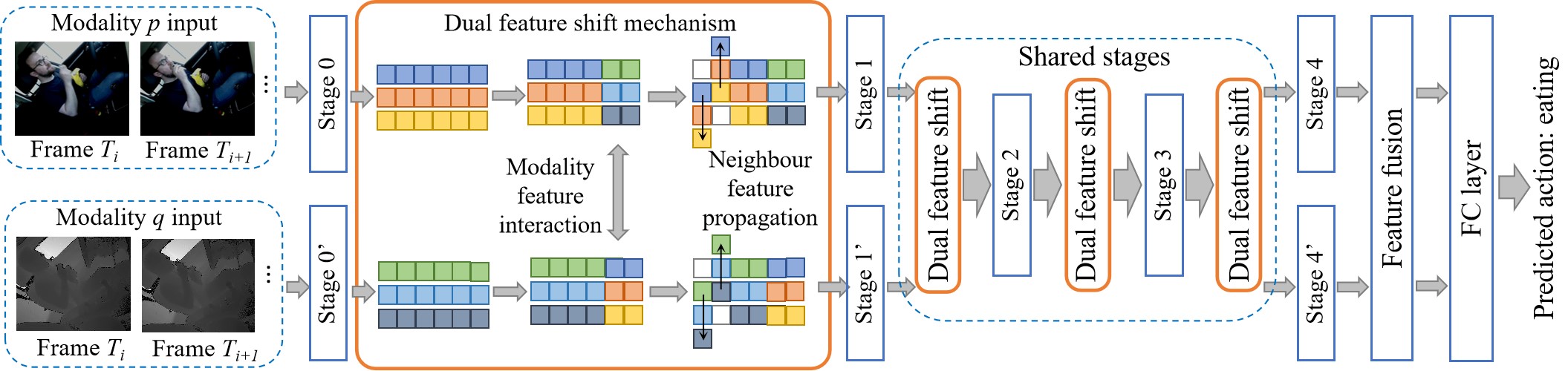}
% \vspace{-3mm}
\caption{Framework of the proposed DFS model. DFS consists of five feature learning stages, followed by the fusion layer and fully connected (FC) layer. Between every two
stages, the dual feature shift mechanism includes both modality and temporal feature interactions. In the middle stages 2 and 3, DFS shares weights among modalities to improve the model efficiency.}
 % At each stage,
 %pl DFS designs
 %pl including modality feature interaction and neighbour feature propagation
\label{framework}
\vspace{-3mm}
\end{figure*}

%pl In real-world scenarios, it is common for vehicle cabins to be equipped with cameras featuring different modalities (such as RGB, IR, and depth) and views (such as front and right top). To this end, the DAR task is inherently multi-modality, with each data modality potentially offering valuable information. Exploring how to effectively leverage multi-modality inputs and extract temporal features is important for improving action recognition in car cabin monitoring.
% positioned at multiple locations

In real-world scenarios, it is common for vehicle cabins to be equipped with cameras that offer different modalities (such as RGB, IR, and depth) and views (such as front and top-right). Consequently, the DAR task is inherently multi-modal, with each data modality potentially providing valuable information. Therefore, investigating effective ways to utilize multimodal inputs and extract temporal features is essential for enhancing DAR in car cabin monitoring.

% a worthwhile and pressing endeavour 
% Multi-modality fusion strategies for action recognition within car cabins have rarely been studied. 
Several methods conducted multi-modality DAR by utilizing score or late fusion strategies.
% \cite{Roitberg2022ACA}.
%  course strategies of 
% Chen et al. designed a DAR system based on the score fusion of optical flow and RGB features
% % where the optical flow and RGB features are learned by two-dimensional CNNs and fused at the last layer 
% \cite{Chen2020DriverBA}. 
Khan et al. utilized average late fusion to detect driving behaviours with depth and IR modalities \cite{DBLP:conf/crv/KhanSSPA22}.
Ma et al. employed a multi-scale channel attention module for the score fusion of depth and IR inputs \cite{DBLP:journals/corr/abs-2210-09441}. 
Alina et al. compared several late fusion strategies for DAR task \cite{DBLP:conf/ivs/RoitbergPMSSS22}.
Jiang et al. built a multi-camera DAR model by training single-camera feature extractors \cite{10226521}.
% However, 
Current methods typically train separate encoders for each modality, which leads to inefficiency in computational complexity. Additionally, these approaches do not adequately consider the temporal correlations among frames.
% reduction in computational efficiency

In this paper, we propose a novel and efficient multi-modality driver action recognition model based on dual feature shift, named DFS. DFS first integrates complementary features across modalities by performing feature interaction along the modality dimension. Then, DFS shifts the features along the temporal dimension within a single modality, thereby facilitating feature propagation between frames. By utilizing both modality and temporal shift operations, DFS can improve inter- and intra-modality feature interactions without additional computational costs. To further learn common patterns and improve model efficiency, DFS shares certain feature encoders among modalities in the framework. To verify the effectiveness of DFS, extensive experiments have been conducted on the Drive\&Act dataset.
% achieve good action recognition performance 
% integrate features across modalities
% These guidelines include complete descriptions of the fonts, spacing, and
%  middle of the feature extraction 
% related information for producing your proceedings manuscripts. Please follow
% them and if you have any questions, direct them to Conference Management
% Services, Inc.: Phone +1-979-846-6800 or email
% to \\\texttt{papers@2021.ieeeicassp.org}.

% \vspace{-2mm}
\section{Methodology}
% \vspace{-2mm}
\subsection{Overview}
For the multi-modality DAR task, the inputs are video clips from $N$ modalities, as $D=\{X^1, X^2, \cdots, X^N\}$. 
For the $p$-th modality, the video clip is $X^p \in R^ {C \times T \times H \times W}$, where $C$ denotes the number of the input feature channels, $T$ denotes the number of the input frames of the clip, and $H$ and $W$ indicate the spatial resolutions of the input feature height and width, respectively. 
%pl The goal of DFS model is to conduct efficient feature fusion for the multi-modality input video clips, to achieve the surpassing performance in driver action recognition.
The objective of the DFS model is to efficiently fuse features from multi-modality input video clips, aiming to achieve superior performance in driver action recognition.

% Our goal is to perform efficient multimodal fusion from unaligned multimodal data sequences,
% in order to obtain the representation that can produce desirable performance in sentiment attitude predictio
% The target of DFS model is to learn the function $f:D $that can map the inputs 

%\subsection{Framework of the proposed DFS model}

The DFS framework is illustrated in Fig.~\ref{framework}, using two modalities as an example. To clarify, the inputs of DFS can be more than two modalities. DFS consists of five stages, followed by the feature fusion layer and a fully connected (FC) layer to generate scores for predicted actions.
We utilize the ResNet \cite{He2015DeepRL} with five stages in total for feature extraction. 
The features at different time stamps are denoted with different colours in each modality.
Between every two stages, the dual feature shift mechanism (detailed in Sec. \ref{DFS}) is employed to integrate complementary features along modality and temporal dimensions.
Also, DFS shares parameters for the middle two stages among modalities to learn common patterns across them.
%pl added "shared parameters"

% \vspace{-2mm}
\subsection{Dual feature shift mechanism}
\label{DFS}
% the beginning of
% At each feature extracting stage, t
The dual feature shift mechanism is designed with the modality feature interaction module and the neighbour feature propagation module. The modality feature interaction module shifts features along the modality dimension, while the neighbour feature propagation module shifts features along the temporal dimension.
% , which shift the features along the modality and temporal dimension, respectively. 
% First, the former module shifts the features along the modality dimension. And the latter one shifts the features along the temporal dimension.
% 

\vspace{-2mm}
\subsubsection{Modality feature interaction}
To learn complementary features from multiple modalities, the dual feature shift mechanism
includes a modality feature interaction module.
% to facilitate feature exchange between modalities.
% Specifically, f
% For the $p$-th modality, the input video clip is $X^p \in R^ {C \times T \times H \times W}$, where $C$ denotes the number of the input feature channels, $T$ denotes the number of the input frames of the clip, and $H$ and $W$ indicate the spatial resolutions of the input feature height and width, respectively.
Modality feature
interaction transfers the feature across different modalities.
For modality $p$ and modality $q$, the $T$-frame video clips are $X^p=\{x^p_t\}_{t=1}^T$, and $X^q=\{x^q_t\}_{t=1}^T$. $x^p_t,\ x^p_t \in R^ {C\times H \times W }$ denote the frames at time stamp $t$.
The feature $x^p_t$ and $x^q_t$ can be updated by shifting the last $k$ feature channels of the modality:
% of modality $p$ can be updated by shifting process from feature $x^q_t$ of modality $q$:
\begin{align}
    \hat{x}^p_t &= M_{shift}(x^p_t,x^q_t)=Concat(x^p_t[:-k], x^q_t[-k:])\\
    \hat{x}^q_t &= M_{shift}(x^q_t,x^p_t)=Concat(x^q_t[:-k], x^p_t[-k:]), 
\end{align}
% where $M_{shift}(\cdot)$ operation exchanges a portion of the feature channels along the modality dimension. 
where $Concat(\cdot)$ denotes the vector concatenation operation via the channel dimension.
This can be conducted with no multiplication cost. As a result, modality feature interaction can integrate additional features across modalities efficiently.

\vspace{-2mm}
\subsubsection{Neighbour feature propagation}
% Inspired by \cite{DBLP:conf/iccv/LinGH19}, 
%pl minor grammatical edits
To leverage temporal correlations among frames, the neighbour feature propagation module further shifts features along the temporal dimension of the single modality frames.
We propagate the information along the temporal dimension in two directions (forward and backwards). For a $T$-frame video clip $\hat{X}^p=\{\hat{x}^p_t\}_{t=1}^T$ from modality $p$, the feature of $\hat{x}^p_t $ can be updated by shifting the first $2i$ feature channels from neighbour frames $\hat{x}^p_{t-1}$ and $\hat{x}^p_{t+1}$: 
% at time stamps $t-1$ and $t+1$ by:
% with $C$ channels and $H \times W $ spatial resolution is the tensor
% $x^p_{t-1}$ and $x^p_{t+1}$ as:
\begin{align}
    \tilde{x}^p_t &= T_{shift}(\hat{x}^p_{t-1},\hat{x}^p_t, \hat{x}^p_{t+1})\\
    &=Concat(\hat{x}^p_{t-1}[:i],\hat{x}^p_{t+1}[i:2i], \hat{x}^p_t[2i:]).
    % \tilde{x}^q_t &= T_{shift}(\hat{x}^q_{t-1},\hat{x}^q_{t+1}), 
\end{align}
This shift operation $T_{shift}(\cdot)$ also has no multiplication cost. Consequently, neighbour feature propagation can improve the intra-modality temporal feature representation efficiently.
% extraction.
% improve the low-latency video recognition.

\vspace{-2mm}
\subsection{Action recognition algorithm}
\vspace{-1mm}
\begin{figure}[t]
% \vspace{2.5cm}
\centering
\includegraphics[width=0.3\textwidth,keepaspectratio]{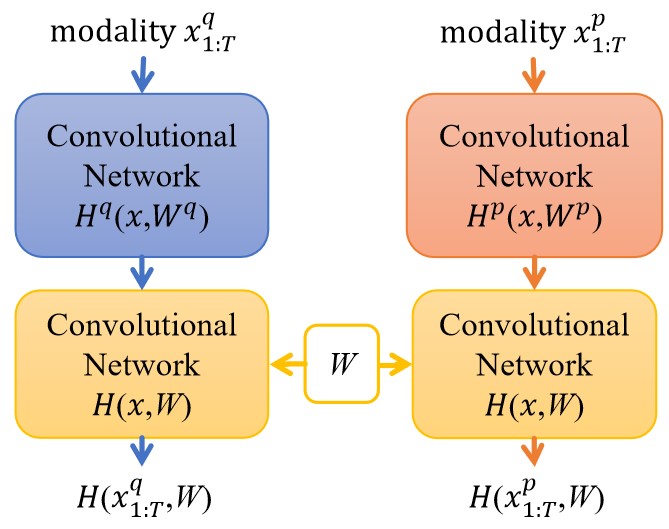}
\vspace{-3mm}
\caption{Illustration of the shared feature encoders among different modalities. The weight $W$ is shared.}
\label{shared}
\vspace{-4mm}
\end{figure}

\iffalse
In Fig.~\ref{framework}, there are five stages during the feature decoder for each modality input. In the 
In the last four stages, we utilize the dual feature shift mechanism before the feature computing and update. The shifted features of video frames $\tilde{X}^p=\{\tilde{x}^p_t\}_{t=1}^T$ are further encoded by a feature extractor $H^p$, i.e.:
\begin{equation}
    f^p=H^p(\tilde{X}^p, W^p), 
\end{equation}
where $H^p(\cdot)$ is the residual convolutional computation at each stage. $W^p$ is the weight matrix to be trained.
\fi

In Fig.~\ref{framework}, DFS consists of five distinct CNN-based stages for feature extraction on each modality. Between every two stages, we incorporate the dual feature shift mechanism, and %prior to the execution of feature computation and subsequent updates. 
the shifted features \( \tilde{X}^p = \{\tilde{x}^p_t\}_{t=1}^T \) of video frames are inputted into the feature extractor \( H^p \) in the next stage. The encoded feature vectors are represented as $f^p = H^p(\tilde{X}^p, W^p)$ with the weight matrix \( W^p \), for modality $p$.

% where \( H^p(\cdot) \) epitomizes the residual convolutional computation undertaken at each stage 

To learn common patterns of modalities and optimize model efficiency, DFS shares the feature encoders for different modalities. As illustrated in Fig.~\ref{shared}, the inputs are initially processed by separate encoders $H^p$ and $H^q$, followed by the shared CNN encoder $H$ with common weight $W$.
% the first two encoder stages are trained for the separate modality. In the middle two stages, the feature encoder stages are the same among different modalities. 
After two shared encoder stages, the features are then updated by the separate encoders and denoted as $\hat{f}^p$ and $\hat{f}^q$.
% last encoder stage is separately trained .
% After five stages of feature learning, the modality-specific spatial features integrated with modality iteration and temporal information can be obtained. 
We then fuse modality features  $\hat{f}^p$ and $\hat{f}^q$ by:
\begin{equation}
    % f=\frac{1}{N}\sum_{p=1}^N f^p
    f_{\text{fusion}}=F(\hat{f}^p, \hat{f}^q, W^F), 
\end{equation}
where $W^F$ is the weight to be trained and the $F(\cdot)$ denotes the feature fusion strategy. The final action predictions can be generated after the FC layer.
% from multiple modalities.
% to form an ensemble feature expression.
% in this paper.
We utilize the cross entropy loss function for model training and optimisation.
 % between the predicted results and the ground truth
% between the predicted results and the ground truth

\vspace{-2mm}
\section{Experiments and results}
\vspace{-2mm}
\subsection{Implementation details}
\vspace{-2mm}
We utilize the pre-trained ResNet-50 \cite{He2015DeepRL} as the spatial feature extraction backbone.
%pl backbone.
% of the feature extraction in each modality, which is initialized with pre-trained parameters on ImageNet \cite{DBLP:conf/cvpr/KarpathyTSLSF14}.
Following \cite{DBLP:conf/iccv/LinGH19}, we sample the input video with a temporal stride 8 and we randomly crop and resize each frame to 224×224.
The $k$ is selected through a grid search, with the optimal performance achieved when k is set to 1/8 of the total channels.
The shared layers are employed in stages 2 and 3 of the feature encoder.
We utilize average pooling to fuse the multi-modality features for the final layer.
%pl We utilize the average fusion to fuse the multi-modality features.
Stochastic gradient descent is used as the optimizer with an initial learning rate of $10^{-4}$. 
% Consistent with previous work \cite{DBLP:conf/iccv/LinGH19, Peng2022TransDARCTD}, 
All experiments are performed on GeForce RTX 3090 GPU.
For performance evaluation, we utilize two commonly used metrics, the Top-1 accuracy (Top-1 Acc.) and the balanced accuracy (Bal Acc.) \cite{DBLP:conf/iccv/MartinRHHRVS19}.
% with a batch size of 8
% or mean class accuracy 

\vspace{-2mm}
\subsection{Dataset}
\vspace{-2mm}
The Drive\&Act \cite{DBLP:conf/iccv/MartinRHHRVS19} dataset is widely used in DAR tasks. It includes 9.6 million frames in three modalities (RGB, IR and depth) and five views. There are three levels of activity labels: action units, fine-grained activities, and coarse tasks. In this paper, we choose the fine-grained level labels with RGB, IR, and depth modalities from the right-top view. We also follow the three-split setting for model training and testing and integrate the results for fair comparisons.

% contains 3 splits for training and evaluation (with no driver overlap between the training, validation and test sets), which we adopt to keep fair comparisons to previous works. The results of the three validation and test sets are averaged. 

\begin{table}[ht]
\caption{The overall results of DFS in comparison with existing methods (Scores in \%).}
\vspace{0.1cm}
\centering
\begin{tabular}[width=0.9\textwidth]{l|cc}
\hline
Methods & Top-1 Acc.& Bal Acc.  \\
% & Top-1           & Top-5           & Top-1         & Top-5          \\
\hline
ResNet \cite{He2015DeepRL}& 56.43& 51.08\\
TSM \cite{DBLP:conf/iccv/LinGH19} & 70.31& 61.11\\
MDBU: Avg fusion \cite{DBLP:conf/ivs/RoitbergPMSSS22} &  74.31                    &   60.25             \\
% Avg. w. weight \cite{DBLP:conf/ivs/RoitbergPMSSS22}            & 59.85                    & 73.79               \\
MDBU: Max Fusion \cite{DBLP:conf/ivs/RoitbergPMSSS22}               &                   72.49 &    59.70                 \\
% Product w/o weight \cite{DBLP:conf/ivs/RoitbergPMSSS22}           & 61.91                      & 76.89                   \\
% MDBU: Product weight \cite{DBLP:conf/ivs/RoitbergPMSSS22}            &                    76.91&   62.02                 \\
% MDBU: Majority      \cite{DBLP:conf/ivs/RoitbergPMSSS22}                &     74.23              &   60.23               \\
% Borda count w/o weight \cite{DBLP:conf/ivs/RoitbergPMSSS22}      & 55.08                      & 70.81                \\
% Borda count w. weight \cite{DBLP:conf/ivs/RoitbergPMSSS22}        & 60.15                    & 74.15               \\
% Reciprocal rank \cite{DBLP:conf/ivs/RoitbergPMSSS22}              & 54.26                    & 70.78                  \\

DFS (Ours)                         & \textbf{77.61}                        &       \textbf{63.12}       \\
\hline
\end{tabular}
\label{baseline}
\vspace{-5mm}
\end{table}

\begin{table}[ht]
\centering
\caption{The comparison results of DFS based on different modalities (Scores in \%).}
\vspace{0.1cm}
\begin{tabular}[width=0.9\textwidth]{l|l|cc}
\hline
% \multirow{2}{*}{} & \multirow{2}{*}{Modality} & \multicolumn{2}{c|}{BalAcc}  &\multicolumn{2}{c}{Top-1} \\
% &  &  Val &  Test &  Val & Test\\
\multicolumn{2}{c|}{Modality setting}  &Top-1 Acc.&Bal Acc.\\
\hline
% \multirow{3}{*}{Single-modality} 
% & Colour       & 68.50      & 62.58          \\
%  & IR       & 69.28      & 60.71          \\
%  & Depth       & 67.43      & 58.03         \\
\multirow{3}{*}{Single} & RGB          & 68.23&62.72       \\
 & IR          & 67.75&59.81         \\
 & Depth        &63.76  &58.28     \\
\hline
% \multirow{4}{*}{DFS(Ours)}      & Colour+IR    &  60.76   & 72.32   \\
 \multirow{3}{*}{Mutiple}      & RGB+IR    & 72.32  &62.87 \\
 & RGB+Depth           &    73.15  & 62.67       \\
& IR+Depth        & 77.61        & 63.12               \\

% & RGB+IR+Depth &    62.79        &    68.94                     \\    
\hline
\end{tabular}
\label{overall}
\vspace{-2mm}
\end{table}
\vspace{-2mm}

% \begin{table}[t]
% \centering
% \caption{Overall results of DFS based on different modalities (Scores in \%).}
% \vspace{0.1cm}
% \begin{tabular}[width=0.9\textwidth]{l|l|ll}
% \hline
% % \multirow{2}{*}{} & \multirow{2}{*}{Modality} & \multicolumn{2}{c|}{BalAcc}  &\multicolumn{2}{c}{Top-1} \\
% % &  &  Val &  Test &  Val & Test\\
% \multicolumn{2}{c|}{Modality} &Bal Acc. &Top-1 Acc.\\
% \hline
% % \multirow{3}{*}{Single-modality} 
% % & Colour       & 68.50      & 62.58          \\
% %  & IR       & 69.28      & 60.71          \\
% %  & Depth       & 67.43      & 58.03         \\
% \multirow{3}{*}{Single} & RGB       &    62.72   & 68.23       \\
%  & IR       &  59.81     & 67.75         \\
%  & Depth       & 58.28      &63.76       \\
% \hline
% % \multirow{4}{*}{DFS(Ours)}      & Colour+IR    &  60.76   & 72.32   \\
%  \multirow{3}{*}{Mutiple}      & RGB+IR    &  62.87   & 72.32   \\
% & IR+Depth    &   63.12        & 77.61                       \\
% & RGB+Depth    &    62.67       &    73.15         \\
% % & RGB+IR+Depth &    62.79        &    68.94                     \\    
% \hline
% \end{tabular}
% \label{overall}
% \vspace{-5mm}
% \end{table}

\vspace{-2mm}
\subsection{Experimental results}
% To evaluate the performance of DFS, w
We design experiments from various angles on the Drive\&Act dataset.
% considering multiple factors, using the Drive\&Act dataset.
%pl .
First, we compare DFS with existing multi-modality DAR models.
% based on multi-modality fusion.
Then, we show the effectiveness of the DFS model with different modality inputs.
% , followed by the ablation study . 
% In addition, we combine the proposed model with different fusion strategies to evaluate the model adaptation. 
In addition, we conduct the ablation study on different feature shift settings to verify the component necessity.
% feature shifts to verify the necessity of each component of DFS.
Finally, the model efficiency and results visualization are analyzed.
% latency and parameters are analyzed to evaluate the model efficiency.

\vspace{-2mm}
\subsubsection{Overall results of DFS on DAR task}
To verify the effectiveness of our DFS model on the DAR task, we compare DFS with ResNet-50 \cite{He2015DeepRL}, and TSM \cite{DBLP:conf/iccv/LinGH19}, multi-modality MDBU modal \cite{DBLP:conf/ivs/RoitbergPMSSS22}.
We reproduce these models using the late fusion strategy on depth and IR modalities.
% Also, we compare with It can be seen that
The results are shown in Table \ref{baseline}. DFS surpasses the existing models on each metric in the table. Specifically, DFS achieves 63.12\% on Bal Acc., surpassing MDBU with average fusion \cite{DBLP:conf/ivs/RoitbergPMSSS22} by 2.87\%.
% (Bal Acc.) and 3.3\% (Top-1 Acc.), respectively.
The results verify the effectiveness of the proposed DFS model in performing multi-modality action recognition, enhanced with the modality feature interaction and neighbour feature propagation.

\begin{table}[ht]
\centering
\caption{The results on different feature shift settings. (M means modality shift and T means temporal shift).}
\vspace{0.1cm}
\begin{tabular}{l|cc}
\hline
% \multirow{2}{*}{Feature Settings} & \multicolumn{2}{c|}{Bal Acc.}                   & \multicolumn{2}{c}{Top-1 Acc.} \\
%  &   \textit{Val}   & \textit{Test}   & \textit{Val}    & \textit{Test}     \\
Feature shifts&Top-1 Acc. (\%)&Bal Acc. (\%)\\
\hline
M+T, shared & 77.61 &  63.12\\
% &  71.30& 72.07
T, shared& 67.73 & 58.03\\
 % &  70.62& 75.37 
T, nonshared &70.31  &  61.11 \\
% & 68.67& 78.39
Nonshift& 56.43  & 51.08\\
% & 58.16& 68.00
\hline
\end{tabular}
\label{AS}
\vspace{-5mm}
\end{table}
% \begin{figure}
%     \centering
%     \includegraphics[width=0.45\textwidth]{bal.jpg}
%     \caption{The balanced accuracy results under different feature shift settings on Drive\&Act.}
%     \label{balacc}
% \end{figure}
\begin{table}[ht]
\centering
\caption{The comparison of the model efficiency results.}
\vspace{0.1cm}
\begin{tabular}{c|l|cc} 
\hline
Modality&Methods                                       & Latency (ms) & \#Param\\
\hline
Single& TSM \cite{DBLP:conf/iccv/LinGH19}& 15.0  &   25.3M    \\
Single&I3D \cite{Wang2018VideosAS} & 18.3   &   28.0M  \\
Single&VST-T \cite{Liu2021VideoST}& 40.2   &   36.5M   \\
Dual &TSM \cite{DBLP:conf/iccv/LinGH19} & 33.0     &  47.2M  \\
% I3D (multi-modality)  & 40.6      & 56.2M  \\
Dual &DFS (ours)            & 28.0      & 38.8M\\
\hline
\end{tabular}
\label{effi}
\vspace{-3mm}
\end{table}

\vspace{-2mm}
\subsubsection{Results of DFS with different modalities} 
% We evaluate DFS with modality combinations including RGB, IR, and depth.
We evaluate DFS on various modality combinations involving RGB, IR, and depth in Table \ref{overall}.
% The depth modality frames 
% capture the distances between the human and the camera and thus 
% can indicate the temporal feature of the actions.
% And the IR and RGB frames can reflect the appearance feature. 
% The overall results with different modality combinations are listed in Table \ref{overall}.
It can be observed that DFS with multiple modality inputs achieves better scores than single modality inputs. DFS with depth and IR as multi-modality input surpasses the single-modality on IR by 3.31\% and on depth by 4.84\%, respectively.
%pl An observation is that there is a limited improvement with RGB and IR modalities as inputs. This can be because the two modalities are similar, and shifting features can supply limited information.
One observation is that there is limited improvement when using RGB and IR modalities as inputs. This can be attributed to the similarity between the two modalities.
% , resulting in limited new information being exchanged during the shift process.
% Another interesting observation is that, when combining all three modalities as input, the DFS performance drops slightly. This can be because that the feature shift among more modalities can be confusing
The overall results verify the performance of integrating the dual feature shift mechanism.
% in the multi-modality DAR task.

\vspace{-2mm}
\subsubsection{Ablation study on different feature shift operations}
To further verify the component necessity in DFS, we conduct an ablation study on different feature shifts. In Table \ref{AS}, four different settings are analyzed: 'M+T, shared' with both modality and temporal feature shifts and shared layers, 'T, shared' with temporal feature shift and shared layers, 'T, nonshared' with temporal feature shift, and 'Nonshift' without feature shift or shared layer. 
% First, the $MT_{shared}$ setting denotes the proposed model with dual feature shifts along both modality and temporal dimensions, with shared layers. Then, the modality shift is excluded, leaving the $T_{shared}$ with only temporal feature shift and shared layers. Furthermore, we remove the shared layers setting, remaining $T_{nonshared}$ setting with only temporal feature shift. Finally, the temporal shift setting is excluded, having $Non$ setting with no dual feature shift or shared layers. 
% The results of the ablation study are demonstrated in Table \ref{AS}. 
We can see that the performance drops when the temporal or modality shift is excluded. In contrast, 'T, nonshared' improves on the scores. In this setting, the model can learn modality-specific features with separate encoders. However, this leads to model inefficiency. 

 % Fig.~\ref{balacc} and \ref{top1acc} intuitively illustrate the detailed score variations on three splits when these feature settings are applied. 
% \begin{figure}
%     \centering
%     \includegraphics[width=0.45\textwidth]{top11.jpg}
%     \caption{The top-1 accuracy results under different feature shift settings on Drive\&Act.}
%     \label{top1acc}
% \end{figure}

% \subsubsection{Model Efficiency Analysis} 
% \vspace{}
\vspace{-2mm}
\subsubsection{Efficiency analysis and results visualization} 
%pl
The model efficiency is crucial for real-time driver monitoring systems. 
% especially in real-world applications. 
% This is due to the limited computation capability of the embedded and edged devices. 
% To this end, 
We further evaluate the model efficiency (latency and parameter size), 
% Specifically, we compare the latency and parameter volume of DFS with existing methods
shown in Table \ref{effi}. We compare with TSM \cite{DBLP:conf/iccv/LinGH19}, I3D \cite{Wang2018VideosAS}, and VST-T (state-of-the-art single-modality DAR) \cite{Liu2021VideoST} in both single- and multi-modality settings.
% Also, we compare with the , a state-of-the-art driver action recognition method on the single-modality setting. 
For latency, DFS improves the processing time to 28.0 ms for one multi-modality input.
%pl one frame input in multi-modality settings.
%pl For parameter volume, DFS consumes less computation than TSM, while achieving surpassing performance on Drive\&Act. <!!-- Less parameters does not mean less computation, in fact the computation is about the same as TSM --!!>
DFS is also lower than TSM for the parameter size while surpassing its performance.
% on Drive\&Act. 
% As a result, DFS is accurate, as well as efficient.

%pl Then we visualize the predicted results for the action 'closing laptop', with different modality inputs in Table \ref{vi}. We can see that the prediction is incorrect with only RGB modality input. But when we combine RGB and depth, DFS gets the correct prediction on this action.
We further visualize some samples of modality inputs and associated results for two actions, ’closing laptop’ and 'eating'. As shown in Table \ref{vi}, the results for single-modality are incorrect (in red).
%pl However, when combining multi-modality inputs, DFS yields the correct results (in green) for the actions by the dual feature shift mechanism.
In contrast, when combining multi-modality inputs, DFS produces the correct results (highlighted in green).
% integrating additional features.
% information
% the ground truth action 
% , with different modality inputs.
% in Table \ref{vi}.
% the prediction is incorrect with single-modality input. 
% It can be seen that 

% \subsubsection{Visualization of the results}
% 对错改成统一格式，对的用绿色，错的用红色或者对的标correct，错的标incorrect
\begin{table}[ht]
\renewcommand{\arraystretch}{0.6}
\centering
\caption{Sample visualizations of the predicted results with different modality inputs (Correct and incorrect results are indicated in \textcolor[RGB]{0,176,80}{Green} and \textcolor{red}{Red}, respectively). }
% \textcolor[RGB]{0,176,80}{Green} result indicate correct prediction, and \textcolor{red}{red} result indicate incorrect prediction.}
\vspace{0.1cm}
\begin{tabular}{m{0.09\textwidth}<{\centering}|m{0.08\textwidth}<{\centering}|m{0.09\textwidth}<{\centering}|m{0.08\textwidth}<{\centering}}
% {l|l|l|l}
\hline
Modality inputs &Results&Modality inputs &Results\\
\hline
&&&\\
\includegraphics[width=0.08\textwidth,keepaspectratio]{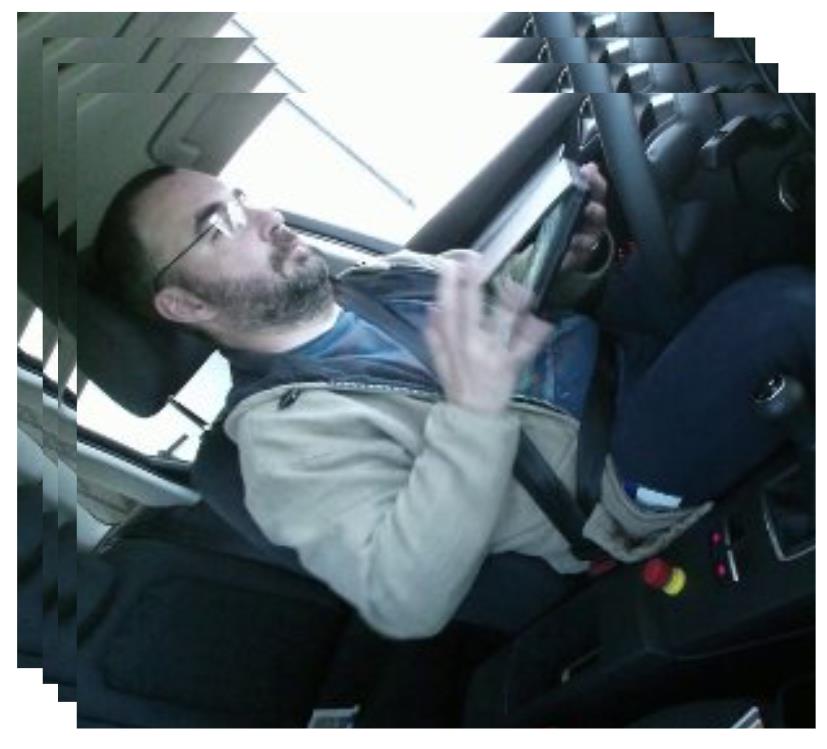} RGB
& \textcolor{red}{opening laptop} & \includegraphics[width=0.08\textwidth,keepaspectratio]{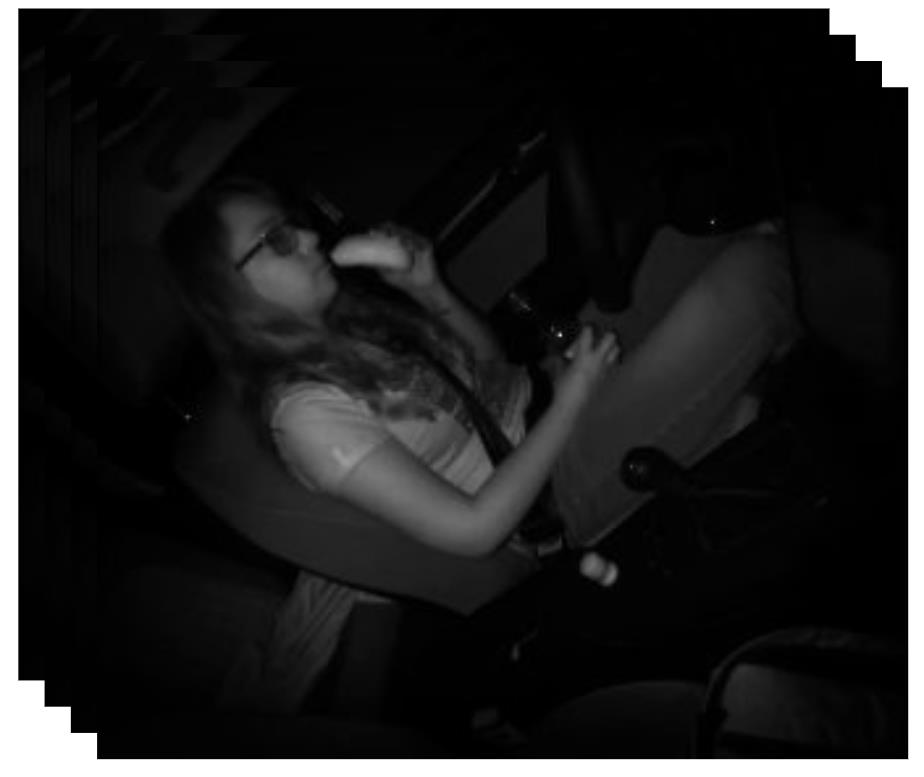} IR& \textcolor{red}{drinking}\\
% \multirow{2}{*}{False}\\
% RGB only\\
\hline
&&&\\
% \vspace{0.1cm}
% \includegraphics[width=0.25\textwidth,keepaspectratio]{depthonly.jpg}& \multirow{2}{*}{True}\\
% Depth only\\
% \hline
% & \multirow{2}{*}{True}\\
% \includegraphics[width=0.25\textwidth,keepaspectratio]{depth1.jpg}  \\
\includegraphics[width=0.08\textwidth,keepaspectratio]{RGB1.jpg}  & \textcolor[RGB]{0,176,80}{closing laptop} & \includegraphics[width=0.08\textwidth,keepaspectratio]{IR2.jpg}  & \textcolor[RGB]{0,176,80}{eating}\\
\includegraphics[width=0.08\textwidth,keepaspectratio]{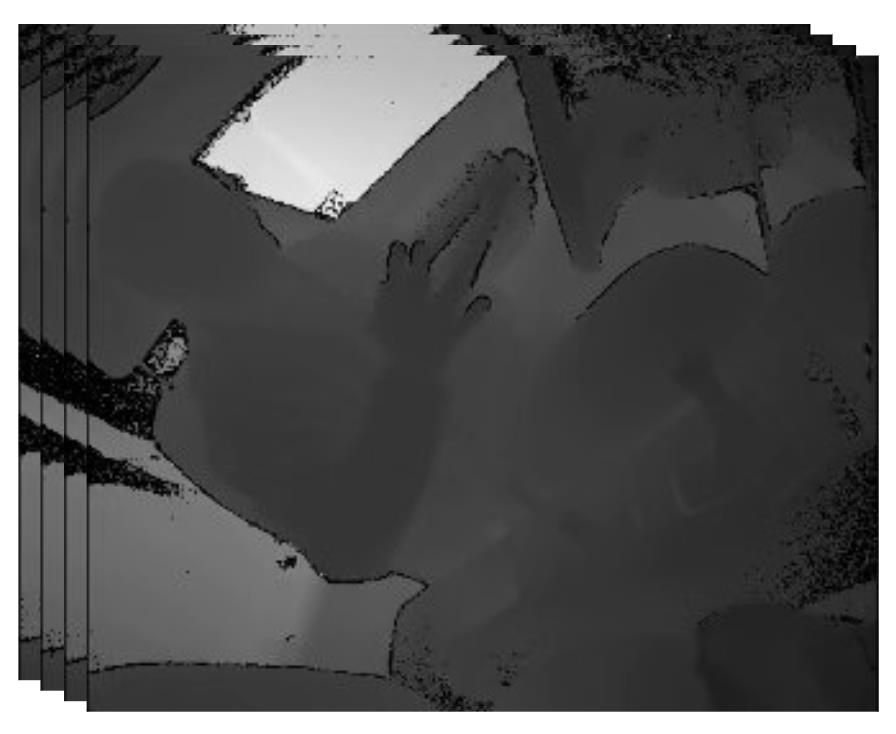} RGB+Depth  & \multicolumn{1}{c|}{}                      & \includegraphics[width=0.08\textwidth,keepaspectratio]{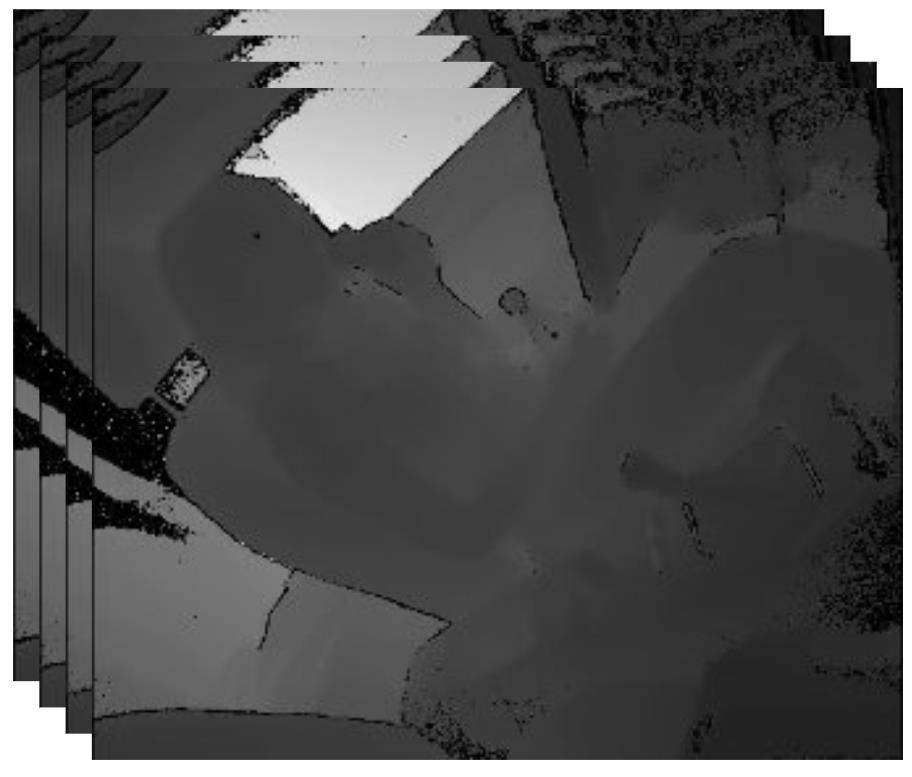} IR+Depth & \\
\hline
\end{tabular}
\label{vi}
\vspace{-4mm}
\end{table}

\vspace{-2mm}
\section{Conclusion}
% we focus on the multi-modality fusion strategies for the driver action recognition task in car cabin monitoring systems. A
In this paper, a novel and efficient dual feature shift model, named DFS, is designed for car cabin monitoring systems. DFS conducts modality feature interactions among different modalities and achieves neighbour feature propagation among single-modality frames. In addition, DFS shares the feature encoder stages among modalities for model training efficiency. Experimental results on the Drive\&Act dataset verify the performance and efficiency of DFS for multi-modality DAR in vehicle cabin monitoring.
\vfill\pagebreak
\bibliographystyle{IEEEbib}
\bibliography{refs}

\end{document}